# Tree Recognition APP of Mount Tai Based on CNN

Zhihao Cao[1], Xinxin Zhang[2]
College of Information Science and Engineering, Shandong Agricultural University, China[1]
The Department of Agriculture of Dongping County, China[2]
czh@sdau.edu.cn, 920775875@qq.com

*Abstract*— **Mount Tai has abundant sunshine, abundant rainfall and favorable climatic conditions, forming dense vegetation with various kinds of trees. In order to make it easier for tourists to understand each tree and experience the culture of Mount Tai, this paper develops an App for tree recognition of Mount Tai based on convolution neural network (CNN), taking advantage of CNN efficient image recognition ability and easy-to-carry characteristics of Android mobile phone. The APP can accurately identify several common trees in Mount Tai, and give a brief introduction for tourists.**

## I. System Development Environment

The system is developed by Android Studio 3.1.3 and TensorFlow1.5. Android Studio is an integrated Android development tool launched by Google. It provides an integrated Android development tool for development and debugging. TensorFlow is one of the most popular libraries for deep learning, which is not only efficient and scalable, but also can run on different devices. The training environment of CNN is Windows 7, and we use CUDA 9.0 and one GTX1080Ti graphics card to accelerate the calculation.

## II. Tree Image Collection

We collected image data in Tianwai Village and Zhongtianmen of Mount Tai and six representative types of trees were selected: cypress, locust, pine, sycamore, ginkgo and magnolia.

The format of images is JPG, and the size is 3024 x 4023, totaling 5000 pieces. Part of the collected image is shown in Figure 1.

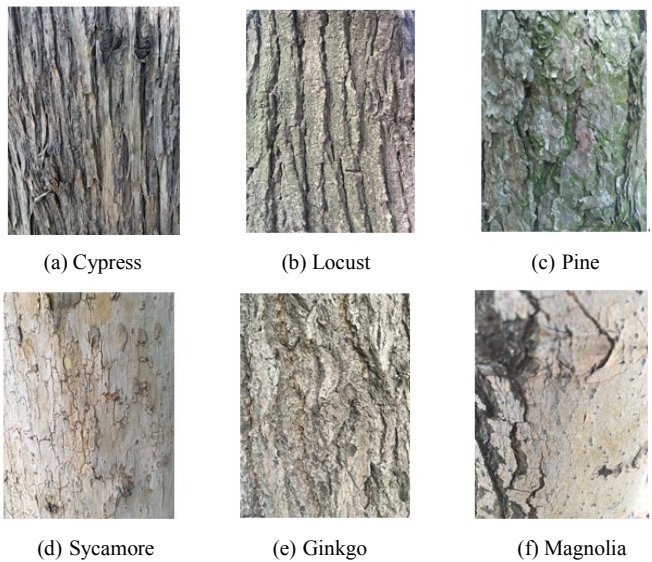

(a) Cypress  (b) Locust  (c) Pine
(d) Sycamore  (e) Ginkgo  (f) Magnolia

Fig. 1. Part of the collected image

## III. System Design

CNN is a common deep learning network model, which is widely used in the field of image recognition. However, the traditional CNN structure is complex and has many network parameters, while the computing power of Android mobile phone is relatively weak. Therefore, the Inception V3 model with less storage space is chosen for the APP. The model structure of Inception V3 is shown in Figure 2.

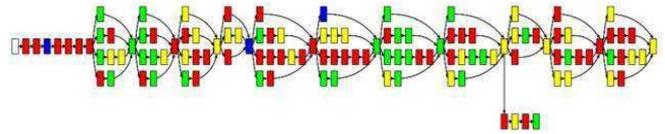

Fig. 2. Model structure of Inception V3

Firstly, Inception V3 network is trained with image dataset, which is carried out on the computer. The corresponding PB model files are generated after training. The generated PB model file is very large, so compression optimization is needed. The optimized PB model file is migrated to Android mobile terminal. The TensorFlow framework is imported into JAVA in the form of JAR packages.

The APP interface is shown in Figure 3. There are two buttons at the bottom of the screen, one is to read the image from the album and the other is to call the mobile camera to take the image. After choosing the image to be recognized, APP will automatically identify the types of trees in the image and give a brief introduction.

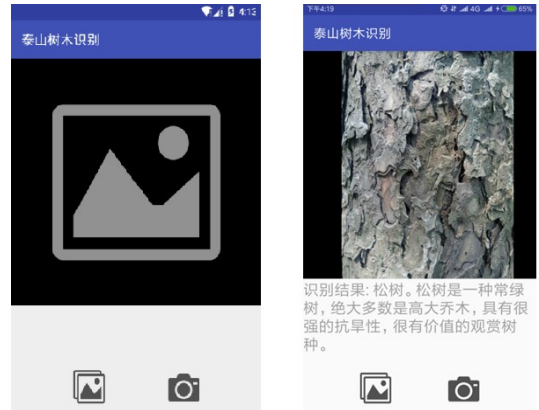

Fig. 3. APP interface

## IV. Source Code Implementation

The source code is divided into two parts: TensorFlow-based CNN source code and JAVA-based Android source code.

Inception V3 network is trained by using Mount Tai tree image dataset. This stage is carried out on the computer. We create a new folder named " data", and then create six folders under the folder, named "cypress", "locust", "pine", "sycamore", "ginkgo" and "magnolia", and then put the training images into the corresponding folder. The CNN code is the Inception V3 source code provided by TensorFlow official *GitHub*. The file name is *retrain.py*.

After training, a model file named "*taishanshumu.pb*" and a text file named "*taishanshumu.txt*" are generated. The PB model file contains CNN network parameters, and the TXT text file contains labels for each category such as "cypress" and so on.

## V. CONCLUSION

We develop a CNN-based tree recognition APP for Mount Tai. This APP can accurately identify several common trees in Mount Tai, and gives a brief introduction to facilitate the use of tourists.